\documentclass{article}

\usepackage{microtype}
\usepackage{graphicx}
\usepackage{subcaption}
\usepackage{booktabs} 

\usepackage{hyperref}
\usepackage{enumitem}

\usepackage[accepted]{icml2026}

\usepackage{amsmath}
\usepackage{amssymb}
\usepackage{mathtools}
\usepackage{amsthm}

\usepackage[capitalize,noabbrev]{cleveref}

\theoremstyle{plain}

\theoremstyle{definition}

\theoremstyle{remark}

\usepackage[textsize=tiny]{todonotes}

\icmltitlerunning{Why Sampling Is Not Choosing}

\begin{document}

\twocolumn[
  \icmltitle{Why Sampling Is Not Choosing: Intentionality, Agency, and Moral Responsibility in Large Language Models}



  \icmlsetsymbol{equal}{*}

  \begin{icmlauthorlist}
    \icmlauthor{Joseph Keshet}{technion}
  \end{icmlauthorlist}
  
\icmlaffiliation{technion}{Faculty of Electrical and Computer Engineering, Technion - Israel Institute of Technology, Haifa, Israel}

\icmlcorrespondingauthor{Joseph Keshet}{jkeshet@technion.ac.il}

  \icmlkeywords{Intentionality, Free will, Normativity, Moral responsibility, Transformer models, LLMs}

  \vskip 0.3in
]



\printAffiliationsAndNotice{}  

\begin{abstract}
Recent advances in large language models (LLMs) have prompted claims that such systems exhibit agency or qualify as moral agents. This paper argues that these attributions are misguided. We maintain that moral responsibility requires commitment-bearing agency grounded in intrinsic intentionality and self-attributed action, and that such agency constitutes the form of free will relevant to responsibility. Although LLMs generate coherent and normatively evaluable outputs, their operation is fully characterized by probabilistic input-output mappings learned from data. Their apparent intentionality is derived rather than intrinsic, and their outputs are neither owned as commitments nor guided by reasons. Variability introduced by stochastic sampling does not amount to choice or authorship. We address objections from the intentional stance, functionalism, compatibilism, and the presence of moral reasoning in model outputs, arguing that none suffice to establish genuine agency.
\end{abstract}

\section{Introduction}

This paper addresses the growing tendency to attribute agency and moral responsibility to LLMs, including multimodal and reasoning models \cite{Zafar2024,Boland2025}. While these systems produce outputs that can be evaluated normatively for correctness, appropriateness, or conformity to moral standards, it remains unclear whether they are appropriate subjects of moral responsibility \cite{Beckers2023,Lima2021}. 

Throughout this paper, we use the term \emph{transformer-based model}, as most LLMs use transformer architectures. However, the arguments developed here apply more broadly to systems trained via next-token prediction and generating outputs through sampling mechanisms. This includes standard LLMs, multimodal systems such as vision–language and speech–language models, and reasoning models built on similar architectures and training paradigms.

The central claim of this paper is that \textit{transformer-based systems, as probabilistic input-output mechanisms lacking intrinsic intentionality and commitment, are not appropriate subjects of moral responsibility, even though their outputs admit normative evaluation.} We support this claim by articulating a set of necessary conditions for moral responsibility in terms of commitment-bearing agency. We argue that such agency requires intrinsic intentionality and the capacity for self-attributed action, and that these together plausibly constitute the form of free will relevant to responsibility. On this basis, we show that transformer-based systems fail to satisfy these conditions.

This work engages debates in the philosophy of AI and mind concerning the agency of artificial systems. Classical arguments by \citet{Searle1980} regarding the syntax-semantics distinction and by \citet{Harnad1990} concerning symbol grounding challenge the possibility of intrinsic intentionality in computational systems. Building on these perspectives, this paper makes two contributions. First, it develops a novel commitment-based account of agency that connects \emph{intrinsic intentionality} to \emph{moral responsibility} through the concepts of \emph{authorship} and \emph{free will}, thereby clarifying the conditions under which a system may be regarded as a genuine bearer of responsibility. Second, it applies this framework to contemporary transformer-based models, offering a systematic analysis of how their probabilistic next-token generation architecture precludes the form of agency required for moral responsibility.

This contribution also aligns with a growing body of recent work that revisits classical critiques in the context of LLMs. In this literature, \citet{Constantinescu2025} examines the social and normative implications of attributing agency to LLMs, while \citet{ai7030099} advances a parallel argument in epistemology. Empirical studies further support this distinction, showing that LLM behavior does not reliably track human-like moral reasoning \cite{doi:10.1073/pnas.2412015122} and is often perceived as only superficially agent-like \cite{10.3389/frai.2025.1710410}. In addition, \citet{zhang2025position} argues that normatively trustworthy systems must satisfy formal verification criteria. Taken together, these works highlight the need to distinguish between systems whose outputs can be normatively assessed and those that genuinely qualify as agents. The present paper contributes to this discussion by providing a unified philosophical and technical account of why transformer-based models fail to satisfy the conditions required for moral responsibility.

\vspace{-.3cm}
\section{Conceptual Framework}

The definitions of intentionality used here are broadly based on \citet{Searle1983}.

\textbf{Intentionality.} Intentionality refers to the characteristic feature of certain mental states whereby they are directed toward, or are about, objects, states of affairs, or propositions. This notion, originating in Brentano's formulation of the ``intentional inexistence'' of mental phenomena \cite{Brentano1874}, and further developed in phenomenology \cite{Husserl1983}, captures the idea that beliefs, desires, perceptions, and judgments are always \textit{of} or \textit{about} something. Intentionality is thus not merely representational structure, but a relation between a subject and a meaningful object as presented to that subject.

\emph{Aboutness} is the core feature of intentionality, referring to the directedness of a state toward an object or content. Whether intentionality is ultimately reducible to physical processes or emerges from them, intentional states are characterized by their aboutness: they are directed toward objects, states of affairs, or contents in ways not captured by ordinary physical descriptions. In paradigmatic cases, aboutness involves reference and meaningful presentation, where the object is given as something (e.g., as threatening, desirable, or true). This aspect is central in phenomenological accounts, where aboutness is inseparable from how objects are disclosed to a subject. The present discussion remains neutral on the metaphysical basis of intentionality; intrinsic intentionality may be physically realized while still differing from merely derived forms of aboutness.

Here we distinguish between two forms of intentionality based on the source of their meaning. \emph{Intrinsic intentionality} refers to a form of aboutness that is inherent to a system itself and does not depend on external interpretation. In such cases, the intentional state has meaning for the subject that entertains it. By contrast, \emph{derived intentionality} refers to a form of aboutness that is not inherent to the system but is attributed to it by external agents, users, or interpretive practices. In these cases, any meaning associated with the system's states is imposed from outside rather than originating within the system itself.

The distinction between intrinsic and derived intentionality can be clarified through contrastive examples. Consider first a human perceptual judgment: upon seeing a dog in a yard, a subject forms the thought, ``There is a dog in the yard.'' This thought is directed toward an object, has determinate content, and is assessable for truth or falsity. If no dog is present, the subject is mistaken. Crucially, the content of the thought is meaningful for the subject itself; the aboutness is internal to the cognitive state. This constitutes a paradigm case of intrinsic intentionality.

By contrast, consider a thermometer displaying ``22°C.'' While the reading appears to be about the ambient temperature, the device does not experience or understand temperature, nor does it entertain a representation that is meaningful for it. If the reading is incorrect, the device is not mistaken but malfunctioning. The apparent aboutness of the thermometer's output depends on external interpretation by users and designers. In this case, intentionality is not intrinsic to the system but derived from human interpretive practices \cite{Searle1980,Harnad1990}.

\textbf{Normativity and Moral Responsibility.} \emph{Normativity} concerns the domain of evaluation in terms of correctness, appropriateness, or obligation. A system or action is normative insofar as it can be assessed relative to standards or rules \cite{Brandom1994}. It is important to distinguish between thin normativity and stronger moral normativity. Thin normativity may apply to any system whose outputs can be evaluated (e.g., correctness of a prediction), whereas stronger forms involve accountability and justification.

\emph{Moral responsibility} is a robust form of normativity involving the attribution of praise, blame, and accountability to an agent. It presupposes that the agent can be understood as the author of its actions in a meaningful sense, capable of endorsing, revising, and being answerable for them \cite{Strawson1962,Williams1985}. Moral responsibility thus requires more than mere causal participation; it involves agency structured by reasons, commitments, and the capacity for self-attribution.

\textbf{Free Will.} Free will, in the sense relevant to moral responsibility, refers to the capacity of an agent to determine its actions in a manner that is not reducible to external causation alone. In existentialist accounts, this includes the ability to transcend given conditions, to select among possibilities, and to commit to a course of action \cite{Sartre1943}. 

\vspace{-.3cm}

\section{The Argument}

The core argument of this paper can be stated as follows:
\begin{enumerate}[nosep]
    \item Moral responsibility requires commitment-bearing agency.
    \item Commitment-bearing agency requires intrinsic intentionality and the capacity for self-attributed action.
    \item Such agency constitutes a form of free will in the sense relevant to moral responsibility.
    \item Transformer-based systems lack intrinsic intentionality and the capacity for self-attributed action, and thus lack free will in the relevant sense.
    \item Therefore, transformer-based systems are not appropriate subjects of moral responsibility.
\end{enumerate}

The first premise states that moral responsibility requires commitment-bearing agency. This reflects a broadly shared view in contemporary philosophy of action: responsibility is not appropriately attributed to mere events or causal processes, but to agents whose actions express commitments, evaluative stances, and responsiveness to norms \cite{FischerRavizza1998,Watson2004}. To hold an agent responsible is to regard its action as something it can answer for, justify, or repudiate. For example, when an individual deliberately deceives another, the act is attributable to the agent as expressing a commitment, however defective, to a course of action and is thus open to blame. By contrast, if a person causes harm during an epileptic seizure, the behavior does not express such commitment-bearing agency and is not typically subject to moral appraisal in the same way. 

The second premise maintains that commitment-bearing agency requires \emph{intrinsic intentionality} and the \emph{capacity for self-attributed action}. Commitment, in the relevant sense, is not exhausted by outward behavior; it involves the agent's capacity to take its actions as its own and to understand them as bearing normative significance \cite{Velleman2000,Bratman2007}. This requires intentional states that are meaningful for the agent itself, rather than merely interpretable by observers. Put differently, commitment-bearing agency depends on intrinsic rather than merely derived intentionality: the relevant aboutness must obtain for the agent, not only from the standpoint of an external interpreter. 

For instance, when a person forms an intention to keep a promise, the intention is directed toward a future action in a way that is intelligible to the agent as something it is committed to doing. The agent can subsequently recognize success or failure relative to that commitment, and can acknowledge or repudiate it as its own. By contrast, systems whose apparent intentionality is merely derived may exhibit behavior that is interpretable as commitment-like, yet lack any standpoint from which such commitments are undertaken or recognized. 

The third premise characterizes \emph{free will} as grounded in such agency. Rather than treating free will as an independent metaphysical capacity, this account understands it as constituted by the kind of agency required for responsibility, namely, the ability to act for reasons, to regard one's actions as one's own, and to exercise a form of control over them \cite{Kane2005,Fischer2006}. On this view, free will does not depend on indeterminism per se, but on reasons-responsiveness and ownership of action. For example, a person who deliberates between competing reasons and acts on what they judge to be the best course of action exercises free will in the sense relevant to responsibility. The emphasis is thus on authorship and rational control, rather than mere unpredictability or absence of causal determination. 

The fourth premise argues that transformer-based systems lack intrinsic intentionality and the capacity for self-attributed action, and thus lack free will in the relevant sense. This claim is grounded in the architectural and functional characteristics of such systems.

Transformer-based models implement a highly structured input–output mapping in which a sequence of input tokens is transformed into a sequence of output tokens through a series of learned parametric operations \cite{Vaswani2017}. At each step of generation, the model computes a probability distribution over possible next tokens using a softmax function applied to internal representations. The final output is produced by sampling from this distribution. Various sampling strategies are used in practice, including greedy decoding (selecting the highest-probability token), temperature scaling (modulating distribution sharpness), top-$k$ sampling (restricting selection to the $k$ most probable tokens), and nucleus (top-$p$) sampling (selecting from the smallest set of tokens whose cumulative probability exceeds a threshold) \cite{Holtzman2020}. These mechanisms introduce variability such that identical prompts may yield different outputs across runs. However, this variability reflects stochastic selection from a learned distribution rather than a process in which the system deliberates, evaluates alternatives, or commits to a particular outcome.

During training, transformer-based models are optimized to predict the next token given preceding context, typically using large corpora of human-generated text. This process employs \emph{teacher forcing}, whereby the model is conditioned on ground-truth previous tokens rather than its own outputs. As a result, it learns to approximate the conditional probability distribution of token sequences without autonomous generation or iterative self-correction.Sampling is not used during training and is applied only at inference. \emph{The objective function does not encode semantic understanding or truth conditions, but statistical regularities in language use}. Consequently, the model's competence arises from pattern recognition rather than any grasp of meaning.

Saussure \citeyearpar{Saussure1916} emphasizes that the relation between signifier and signified is arbitrary and sustained by differences within a system of signs rather than by intrinsic connections to external reality. Transformer models operate precisely at the level of such differential structures, capturing statistical relations among tokens without access to referents or meanings beyond the system. In this sense, their outputs exemplify a purely relational form of sign manipulation, consistent with derived but not intrinsic intentionality.

\citet{Searle1980} argues that the manipulation of formal symbols according to syntactic rules is insufficient for genuine understanding or intrinsic intentionality. Transformer systems instantiate this distinction: they process and generate symbol sequences according to learned statistical rules, yet there is no evidence that these processes are accompanied by semantic comprehension or meaningful intentional states for the system itself. 

Finally, \citet{Harnad1990} formulates the symbol grounding problem to highlight the difficulty of explaining how purely symbolic systems can acquire meaning that is not parasitic on external interpretation. Transformer models, trained exclusively on textual data, operate within a closed symbolic domain where tokens are defined only in relation to other tokens. Without grounding in sensorimotor or experiential interaction with the world, there is no mechanism by which meanings could become intrinsic to the system.

This concern extends, albeit in a modified form, to multimodal LLMs, which integrate textual processing with visual or auditory encoders to produce context-sensitive outputs grounded in multiple input modalities. Empirical studies suggest that their grounding remains limited, and often exhibit hallucinations arising from imbalances in modality integration, where textual representations dominate the generation process while visual or auditory inputs are underutilized \cite{liu2024paying,zou2024look,hsu2025reducing,asadi2026mirage}. As a result, even when non-textual inputs are present, the model's outputs reflect predominantly internal statistical associations rather than genuinely grounded semantic content, reinforcing the claim that their apparent intentionality remains derived rather than intrinsic.

Closely related to the absence of intrinsic intentionality is the absence of self-attributed action. Commitment-bearing agency requires that an agent be able to take its actions as its own, recognizing them as expressions of its commitments and being answerable for them. Transformer-based systems lack this capacity. They do not possess a unified perspective from which outputs are generated as endorsed or disavowed. Instead, each output is the result of a computation that does not involve self-reference or ownership.

The role of probabilistic sampling further underscores this point. Transformer systems employ stochastic mechanisms such that identical prompts may yield different outputs across runs. While this variability might superficially resemble choice among alternatives, it does not amount to agency or free will. Stochastic generation should be distinguished from agency: randomness introduces indeterminacy, not authorship; output generation does not involve endorsement or commitment; and there is no evidence of a first-person perspective or self-attribution of action. Crucially, \emph{free will is not equivalent to randomness or indeterminacy; it involves authorship and responsibility}. An action is free in this sense not because it is unpredictable, but because it is attributable to the agent as its own. 

This characterization extends to \emph{reasoning models}, which augment standard generation with intermediate steps such as chain-of-thought prompting, self-consistency, or tree-based search over candidate outputs \cite{Wei2022,Wang2023,Yao2023}. While such methods may give rise to outputs that resemble structured reasoning by generating multiple candidate continuations, exploring alternative solution paths, or selecting among them according to heuristic criteria, they do not thereby instantiate deliberation in the relevant sense. The generation of intermediate steps remains governed by the same underlying mechanism: the production of token sequences according to learned statistical patterns. Likewise, the selection among alternatives is typically based on likelihood, consistency, or externally defined scoring procedures, rather than on the recognition and evaluation of reasons as reasons. For example, a model may produce several candidate explanations for a problem and select the most frequent or internally consistent one, yet it does not thereby understand those alternatives as competing reasons or endorse one as its own conclusion. The resulting behavior may simulate aspects of reasoning, but it does not involve authorship, commitment, or reasons-responsive control. Accordingly, even in reasoning-augmented models, the processes underlying output generation remain forms of statistical selection rather than genuine deliberation.

These features jointly indicate that transformer-based systems lack the structural conditions required for free will in the sense relevant to moral responsibility. In the absence of intrinsic intentionality and self-attributed action, there is no basis for authorship, commitment, or reasons-responsive control. Accordingly, while such systems generate outputs that are open to normative evaluation, they do not qualify as agents capable of free action.

In Appendix A, we address objections from the intentional stance, functionalism, compatibilism, and the presence of moral reasoning in model outputs.

\vspace{-.3cm}
\section{Conclusion}
\vspace{-.2cm}

This paper has argued that, although transformer-based systems produce outputs that can be normatively evaluated, they are not appropriate subjects of moral responsibility. The implications are both philosophical and practical. Responsibility for these systems remains with human designers, deployers, users, and institutions, rather than with the systems themselves. Accordingly, efforts to align or regulate transformer-based systems should not be understood as transforming them into moral agents, but as mechanisms for constraining and guiding their behavior within human normative practices. Likewise, while such systems may assist human decision-making, they should not be regarded as bearers of moral judgment or responsibility.

\bibliography{main}
\bibliographystyle{icml2026}

\appendix
\section{Objections and Replies}

\textbf{Objection 1: The Intentional Stance.}  
It may be argued, following \citet{Dennett1987}, that if a system can be fruitfully treated as having beliefs and desires, this suffices for attributing agency.

\textit{Reply.} Such attributions are predictive and pragmatic rather than ontological. The intentional stance explains how we model and anticipate behavior, but it does not establish that the system possesses intrinsic intentionality or commitment-bearing agency. Treating a system \emph{as if} it has beliefs is not equivalent to those beliefs being meaningful for the system itself, nor does it ground the authorship required for moral responsibility.

\textbf{Objection 2: Functionalism.}  
A functionalist might argue that mental states are individuated by their causal roles rather than by their physical realization, and that any system exhibiting the appropriate functional organization thereby instantiates the relevant mental properties \cite{Putnam1967,Fodor1975}.

\textit{Reply.} Even if transformer systems realize complex functional structures, the present argument targets the absence of intrinsic intentionality and self-attributed action. Functional equivalence at the level of input–output behavior does not by itself establish that the system's states are meaningful for it or that its outputs are owned as commitments. Without such intrinsic significance and authorship, functional organization alone is insufficient to ground moral responsibility.

\textbf{Objection 3: Compatibilism.}  
A compatibilist might argue that free will does not require indeterminism, but rather that actions be appropriately caused, typically through mechanisms that are responsive to reasons \cite{FischerRavizza1998}.

\textit{Reply.} The present account is compatible with this view. However, even under compatibilism, free will requires that an agent be responsive to reasons and capable of regarding its actions as its own. Transformer-based systems do not evaluate reasons, recognize them as such, or endorse their outputs as commitments. Their behavior is governed by statistical correlations rather than reasons-responsive control, and thus fails to satisfy even compatibilist conditions for agency.

\textbf{Objection 4: Moral Reasoning Outputs.}  
Transformer systems are capable of generating sophisticated moral arguments, engaging with ethical dilemmas, and producing contextually appropriate normative judgments. It may therefore be argued that such systems exhibit the kind of reasoning characteristic of moral agents.

\textit{Reply.} The production of moral language does not entail the possession of moral agency. While transformer systems can simulate reasoning by reproducing learned linguistic patterns, they do not take a stance toward the claims they generate, nor do they recognize those claims as binding. For example, a model may assert that “one ought to keep promises” in one context and later generate a justification for breaking promises in another, without any capacity to reconcile these outputs or to regard either as authoritative. The appearance of reasoning is thus a product of statistical pattern generation rather than the exercise of judgment. Simulation of reasoning does not entail commitment, and without commitment there is no basis for attributing moral responsibility.

\end{document}